# Automated Generation of High-Quality Medical Simulation Scenarios Through Integration of Semi-Structured Data and Large Language Models

**Scott Sumpter CHSOS**
Canadian Surgical Technology and Advanced Robotics
London Health Sciences Centre
scott.sumpter@lhsc.on.ca

## Abstract

Introduction: The advancement of medical education increasingly relies on the use of simulation scenarios to improve clinical skills and patient care. Traditional development of these scenarios is time-intensive and lacks the flexibility to quickly adapt to learning needs. This study proposes an innovative approach to address these challenges.

Methods: A novel framework was developed that combines semi-structured data requests with Large Language Models (LLMs), primarily OpenAI's ChatGPT3.5, to automate the generation of medical simulation scenarios. This methodology uses the capabilities of artificial intelligence to produce detailed, content efficiently. By integrating semi-structured data, the system guides LLMs to create scenarios tailored to specific learning objectives, streamlining the scenario development process.

Results: The use of this AI-enhanced framework has led to a significant reduction in scenario development time and resources, enabling the production of a wider variety of simulation scenarios. Early feedback from educators and learners indicates improved engagement illustrating this framework's effectiveness in enhancing simulation-based learning.

Conclusion: The integration of semi-structured data requests with LLMs presents a transformative approach to scenario generation in medical education. These findings highlight the potential of this methodology to greatly improve the development and application of simulation scenarios. This approach underscores the importance of leveraging AI to improve learning outcomes and patient care standards.

## Introduction

The development of medical simulation scenarios is critical in healthcare education, offering a safe and controlled environment for professionals to practice and refine their skills. These simulations are vital for preparing medical personnel to effectively respond to real-life situations, enhancing patient care outcomes.[1] Traditionally, the creation of these scenarios has been a manual and time-intensive process, limiting the ability of educational programs to offer a varied and complete training experience.

Experiential learning, essential for enhancing technical skills, decision-making, and teamwork in medical education, benefits from the use of high-quality clinical scenarios. The importance of simulation in medical curricula illustrates its effectiveness in bridging the gap between theory and practical application.[2] The cost-effectiveness of simulation-based learning, despite the initial investment, demonstrates its ability to enhance clinical competence and improve patient safety by reducing medical errors.[2]

Further exploration into the role of simulation in team-based learning stresses the need for scenarios that are realistic, practical, and comprehensive. The discussion around the careful attention to detail required by scenario writers in customizing simulations for interdisciplinary team training showcases the creation of scenarios that accurately mimic real-life clinical settings, including event triggers and environmental distractors.[3]

Recent advancements in artificial intelligence, particularly large language models (LLMs), have been explored as a promising solution to these challenges. By leveraging AI to automate the generation of detailed, realistic scenarios, there is potential to significantly reduce the time and effort required to create effective training materials.[4] Discussion with educators and simulation developers reveals that creating these scenarios manually can take anywhere from an hour to over eight hours, depending on the complexity of the scenario and the learning objectives. This considerable investment of time underscores the need for more efficient methods. The automation offered by LLMs, and similar AI technologies promises to dramatically streamline this process, potentially reducing scenario development time to a fraction of what is currently required, thereby enabling a more dynamic, targeted and responsive approach to medical education. However, integrating LLMs into the scenario generation process is not without its challenges, including ensuring clinical accuracy, relevance, and educational value of the generated content.[4]

This work introduces a novel method that significantly reduces the manual workload involved in scenario creation by using semi-structured data to precisely guide the information generation process. This approach not only automates the bulk of the scenario development process but also uniquely parameterizes the data, enhancing its utility beyond mere scenario creation. The parameterization of data allows for its application in a wide array of uses outside traditional scenario development, such as analytical studies, learning content customization, and the development of adaptive learning systems. This versatility underscores the potential of this method to serve as a foundational tool in medical education, enabling not just the creation of high-quality simulation scenarios but also supporting a broader spectrum of educational innovations.

While the automation provided by this method promises to dramatically streamline the scenario development process, it does not eliminate the need for a subject matter expert (SME) to review and possibly correct the generated content to ensure clinical accuracy and relevance. Acknowledging the critical role of SMEs, this approach facilitates a more efficient workflow where the primary source of effort is automated, allowing SMEs to focus on the critical aspects of content verification and educational value.

Unlike the predominantly manual methods currently in use, this approach offers a scalable, efficient solution for medical education, potentially transforming how simulation scenarios are developed and utilized. This method not only enhances the adaptability and realism of simulations but also allows for rapid iteration and customization to meet specific training needs, marking a significant step forward in the field of medical education by improving the efficiency of simulation-based learning and broadening the scope of its application.

# Methods

This methodology introduces a novel approach to generating medical simulation scenarios by leveraging semi-structured data and transformer models. This process not only streamlines the creation of scenarios but also ensures their relevance and adaptability to various medical specialties. Below, the components of this methodology are detailed:

### Semi-Structured Data Approach

The foundation of this method lies in the design of semi-structured data templates, which serve as blueprints for the scenarios. These templates are designed to capture the essential elements of medical simulation scenarios, including patient information, clinical settings, medical conditions, diagnostic procedures, and treatment options. The design process involves:

### Identification of Key Components

In developing these scenarios, the development involved collaboration with medical educators and experts in simulations to figure out the key parts needed for a full and useful scenario. This includes details about the patient, where the scenario takes place, what medical issues are involved, the steps for diagnosing these issues, and the options for treatment.

It's crucial to think about every part of the simulation. Doing so not only helps later but also makes sure the AI properly interprets the context better, leading to better results while keeping within what the AI can do. This covers everything from starting the simulation, giving instructions, to wrapping up with a debrief.

### Template Structuring

In the development of this data template, a multifaceted approach is essential to cater to both the technological nuances of artificial intelligence (AI) and the practical applications of the generated data. The design of this template is carefully planned to address several key considerations:

Understanding and accommodating AI-specific limitations, such as the context window (the amount of data AI can consider at one time), token limits (the maximum number of words or characters the AI can process in a single instance), and the propensity for hallucinations (instances where AI generates incorrect or nonsensical information), are paramount. These constraints necessitate a template that maximizes information density while minimizing the risk of overloading the AI or encouraging inaccurate responses.

To help with efficient data extraction and utilization in scripts or software programs, the template is structured hierarchically. This organizational method ensures that data is grouped in a way that is understandable to the human as well as the AI. Hierarchical structuring aligns with programming practice as well as with AI interpretation considerations.

An example of this can be seen in Table 1: Example of General Info Section Components and Organization. The elements are organized in a way that facilitates an organized and easily referenced format.

### Aligning with AI Interpretation Processes

There's an intrinsic aspect of designing data templates that aligns with how AI processes information, a quality that becomes apparent through experience. This involves crafting prompts and data fields in a way that naturally guides the AI towards generating coherent and contextually appropriate responses. Understanding the AI's tendencies, such as how it links concepts or interprets instructions, is critical. This insight is developed over time, through trial and error, and by analyzing the AI's output patterns in response to various data inputs and structures.

As seen in Table 1 and Table 2, the data has been separated to establish context and facilitate the AI's ability to interpret the input. The data templates for these elements are processed separately but within the same LLM conversation, or context. The first establishes the scenario framework and general information while the second covers the specific progression of the scenario. In essence, the plot is being established and the script scenes are being written.

| Section: General Info | | | |
|---|---|---|---|
| Subsection | Attribute | Type | Description |
| Sim Info | Author | string | The author of the scenario. |
| | Facility | string | The facility where the scenario is based, e.g., "London Health Sciences Centre". |
| | Scenario Title | string | The title of the scenario. |
| | Scenario Description | string | A detailed description of the scenario. |
| | Simulation Objective | string | The main objective of the simulation. |
| | Target Audience | string | The intended audience for the scenario. |
| Case Summary | - | string | A brief summary of the patient's condition and scenario context. |
| Learning Objectives | Objective | array | The specific knowledge or skills to be acquired from the scenario. |
| Equipment Props | Equipment | array | A list of all equipment necessary for the scenario. |
| Environment | - | string | The setting of the scenario, including surroundings and people present. |
| Case Presentation | - | string | Initial presentation of the patient, including clinical details and background information. |
| Debriefing Points | Point | array | Key discussion points for debriefing to enhance learning outcomes. |
| Lab Results | Test | array | Details of possible lab tests, results, and normal ranges. |

*Table 1 General Info Components Organization*

| Section: Scenario Progression | | | |
|---|---|---|---|
| Subsection | Attribute | Type | Description |
| Phase | Id | number | The identifier of each phase in the scenario progression. |
| | Title | string | The name or title of the phase. |
| | Vitals | object | Detailed vital signs to monitor during the phase, including BP, HR, RR, SpO2, Temp, Rhythm, GCS, and Other. |
| | StateModifiers | object | Actions or modifiers that impact the patient's condition, with results. |
| | Learner Actions | object | Actions or steps to be taken by the learners during the phase. |
| | Transition Triggers | array | Conditions for transitioning to the next phase based on learner actions. |

*Table 2 Scenario Progression Components Organization*

**Transformer Integration**

To guide the model in a way that the desired output is achieved, communication directly with the LLM occurs at two levels.

The first level is an overarching prompt that precedes the JSON portion of the template. The purpose of this prompt is to provide the model with the specific requirements of the output as well as how it should utilize the JSON itself.

The second level of prompt is embedded directly within the data template. Due to the utilization of JSON to create the template, we can use description tags to provide specific guidance as to what is desired in each individual section or part of the hierarchy, ensuring the generated content closely adheres to our detailed requirements. Most semi-structured data formats offer similar functionalities, so this two tier method can be utilized across different data formats.

An attempt was made at providing examples in the template to help guide the model in the position in the hierarchy that would be expected in the output. Examples were found to help guide the model but were also found to introduce potential bias. For example, in one scenario about a brain bleed, there was example text for a myocardial infarction. The output added myocardial infarction symptoms to the

Employing a dual-level approach to prompt design effectively directs the AI's content generation process, ensuring it produces information that is both relevant and aligned with educational and clinical objectives.

**Software Integration**

The implementation of this process is tailored to specific requirements. For our purposes, a form using C# with WPF was developed as illustrated in Figure A1. This form includes an input field for educators to specify the subject or details of the simulation they wish to create.

This is then added to the initial prompt in a way that creates a complete instructional prompt to guide the creation of the scenario. The data template is then appended after the initial prompt to create the entire request to bring to the LLM.

Once the model fills out the data template, the data is deserialized for application use. Considering the LLMs potential for misinterpretations or unexpected outputs, incorporating a validation step for the data template is essential. This process flow, illustrated in Figure 2: Process Flow Diagram, outlines the sequence: compiling the information package, submitting it to the model, validating the returned data, fixing any format errors, and finally, presenting the data to the user for approval. Users can either accept the generated data, adjust their prompts for further refinement, or continue refining by further prompts to better define the generated data. The approved data then sets the stage for the next phase (e.g., scenario progression), which undergoes the same steps.

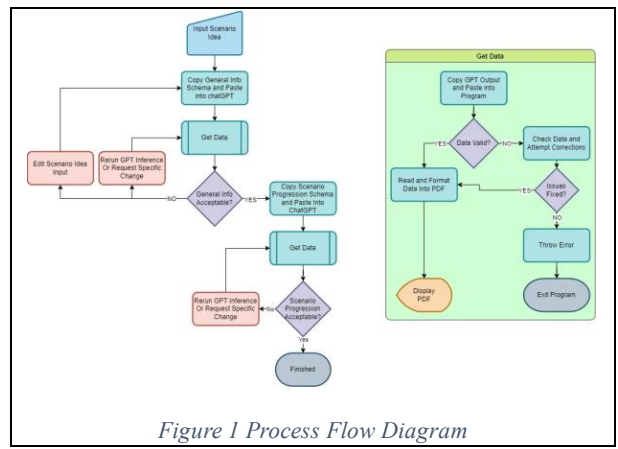

*Figure 1 Process Flow Diagram*

Once the data has been verified, the software can then output the finished scenario in a form that best suits the needs of the simulation center. Figure A2 shows a PDF output created from the data programmatically through the software.

A secondary use for the data as illustrated in Figure A3, is using the data to prepopulate the programming phases of a manikin-based patient simulator. It uses the values returned in the scenario progression to populate the appropriate fields and output in the format used by the simulator.

**Quality Control**

To guarantee medical accuracy, relevance, and educational value, generated scenarios are carefully reviewed and refined as necessary. This review process involves critical evaluation by subject matter experts who ensure that each scenario aligns with current clinical standards and educational goals, adjusting to optimize realism and learning outcomes. Generally, this is performed by the educator who initially requested the scenario as they're likely most familiar with the desired learning outcomes.

### Output Generation

The populated templates are transformed into readable outputs, such as PDF, Word or HTML documents by using widely available open-source software libraries. Custom styling and visual aids enhance readability and usability for educators and learners.

### Choice of Platform

Given the closed source nature and rapid development of the online AI chatbot platforms, it's difficult to define the best model to use for this input. The chosen type of platform was made based on accessibility to a larger number of people, so a commercially available platform would enable this without programming or in-depth AI knowledge.

To preface, all uncited observations below are purely anecdotal and as of the time of this writing. Initial tests were performed across multiple platforms, including OpenAI's ChatGPT-3.5 (GPT3.5) and ChatGPT-4 (GPT4), Google Bard (now known as Google Gemini), Claude, and Microsoft Bing (which is powered by GPT4 API). At the time, multiple issues were observed across all platforms outside of those offered by OpenAI.

The most common issue was output getting cut off before finishing, and going into a loop where the same words would repeat continuously. This may have been caused by an issue with how the prompts were being structured, but no adjustments made eliminated these issues. Currently, the platform most used for this purpose is GPT3.5. GPT4 is admittedly a more accurate and robust model[5] but the significantly faster speed of GPT3.5[6] enabled for faster development iterations when creating the scenario, which was preferable to the slower but potentially more accurate GPT4.

It was also observed that GPT4 would cut off output more often or produce an error before finishing. One interesting observation made that effected this choice, is that during the scenario progression phase, GPT4 would often produce less scenario steps than GPT3.5. When asked to produce 3-6 steps, GPT4 would only ever produce 3 upon first iteration, regardless of the complexity of the scenario, while GPT3.5 would often produce more steps depending on the complexity of the case. This effect was never quantified during the development process and may have changed with newer versions and updates of GPT4.

### Streamlined Example

To effectively demonstrate this approach, a streamlined example will be presented. This involves the strategic organization of essential components into a structured hierarchy, optimized for both efficient data utilization and adherence to the constraints imposed by large language models.

### Identification of Key Components

Extracted from current literature and existing simulations, key components of the simulation creation process are first identified and then organized into a hierarchy that will best suit both the data use and manipulation, as well as operating within the constraints of LLMs:

Scenario

    1.1. General Info
        1.1.1. Case Presentation
        1.1.2. Scenario Title
        1.1.3. Case Summary
        1.1.4. Debriefing Points
        1.1.5. Learning Objectives
2. Scenario Progression
    2.1. Steps
        2.1.1. Step Number
        2.1.2. Patient Status
        2.1.3. Vitals
        2.1.4. HR
        2.1.5. BP
        2.1.6. SPO2

The above example hierarchy only contains a small portion of the general information, and only the minimum progression data. Observing the structure, it becomes evident that the information has been delineated into two primary sections: 1. General Information, and 2. Scenario Progression. The rationale behind this division unfolds as follows:

### Establishing an Overview (General Information)

The first step involves creating a comprehensive overview of the scenario, encapsulated under the "General" section. This includes critical details such as Patient Information (Age, Gender, History) and Case Information (Case STEM, Learning Objectives, Starting Vitals).

This overview serves a dual purpose: it allows for an initial assessment to ensure the scenario generation is on the right track and establishes a foundation for the detailed scenario progression.

### Generating Scenario Progression

The second step focuses on the dynamic development of the scenario, detailed under "Scenario Progression." This section outlines the steps of the scenario, including Step Number, Patient Status, and Vitals (HR, BP, SpO2), reflecting how the scenario unfolds over time.

The separation into two distinct steps optimizes the use of token budgets and maintains context, striking a balance between simplicity and the depth of scenario development.

### Iterative Development and Adaptation

The layout and organization of scenario data were developed through an iterative process, informed by practical experimentation with what works within the constraints of chatbot platforms. This approach facilitated the refinement of our methodology, ensuring efficient data retrieval and processing while navigating the complexities of token limits and context preservation.

By dividing the scenario creation into an initial overview and detailed progression, a methodology was achieved that balances simplicity with the need to maintain context and stay within token budgets. This division enhances the manageability of the scenario generation process and ensures that the resulting scenarios are detailed and aligned with the intended educational outcomes.

*Note, for the rest of the worked example, only the General information will be represented for brevity, Appendix B will have the full example.*

### Template Structuring

Organizing these elements into a JSON structure allows for easy manipulation and data entry. The following JSON schema illustrates how the above components would be used to create a template.

```json
{ "Scenario": {
    "type": "object",
    "properties": {
      "GeneralInfo": {
        "$id": "#root/Scenario/GeneralInfo",
        "type": "object",
        "properties": {
          "CasePresentation": {
            "$id": "#root/Scenario/GeneralInfo/CasePresentation",
            "description": "The initial presentation of the patient in the scenario, including relevant clinical details. Should include a readout of key vitals as well as if any tests were performed or any other inbound issues occurred. Medical background should be provided if it would be available.",
            "type": "string"
          },
          "ScenarioTitle": {
            "$id": "#root/Scenario/GeneralInfo/ScenarioTitle",
            "type": "string"
          },
```
JSON CONTINUES, REFER TO APPENDIX B FOR FULL EXAMPLE…

This architecture not only acts as a foundational framework for the scenarios but also directs the model towards producing content that best fits the needs of the scenarios creator. The following sections detail the essential elements of the JSON template and their role in enabling LLM-driven content generation.

As previously mentioned, "Description" is a pivotal field as it acts as an in-template prompt for the LLM. The descriptions provided in various sections of the JSON structure guide the LLM on what content needs to be generated for each part of the scenario. By clearly outlining the expected content, such as the patient's initial presentation or a brief scenario summary, the descriptions help in directing the LLM's output towards the desired information, thereby enhancing the relevance and accuracy of the generated content.

"Data Type" and "ID" play a crucial role in structuring the JSON template and guiding the LLM. The data type (e.g., string, array, object) informs the LLM of the expected format of the generated content. The ID, including the path, provides a unique reference to each element within the JSON structure. This level of detail is essential for ensuring that the LLM generates content not only contextually appropriate but also correctly formatted and placed within the overall scenario's structure.

### Prompt generation:

This is what will be used to relay the overall wishes to ChatGPT or other LLM. In practice, simply having the software paste something like the following at the beginning has shown the best results:

*Adhering strictly to the following schema, fill in the values to create a complete medical simulation. Output should be the JSON object without any schema elements. The simulation will be about a INSERT SPECIFIC ISSUE HERE:*

General prompt engineering good practices are critically important here.

For a myocardial infarction, the output to paste into the LLM might be:

```
Adhering strictly to the following schema, fill in the values to create a complete medical simulation. Output should be the JSON object without any schema elements. The simulation will be about a myocardial infarction:{ "title": "Scenario",
    "type": "object",
    "description": "", …. Continuation of schema.
```

After pasting this in the ChatGPT input box, it produced the following output:
```json
{
  "Scenario": {
    "GeneralInfo": {
      "CasePresentation": "A 58-year-old male presents to the emergency department complaining of chest pain radiating to his left arm. He appears diaphoretic and anxious. Initial vital signs are as follows: HR 110 bpm, BP 160/100 mmHg, RR 22 bpm, SpO2 96% on room air. An ECG is performed showing ST-segment elevation in leads II, III, and aVF, consistent with inferior myocardial infarction.",
      "ScenarioTitle": "Myocardial Infarction Simulation",
```
JSON CONTINUES, REFER TO APPENDIX B FOR FULL EXAMPLE…

Appendix B contains an example of a simple python script that parses and displays the returned data. The output of the python function can be seen in Table B1: Python Code Output for Example JSON. This took ChatGPT approximately 3 seconds to output and has produced a human readable medical simulation scenario specific to the input that was provided to it. At this point, the user can type in the ChatGPT input box any further refinements, and it will return an adjusted JSON output. For example, typing in "Make the patient a young female" alters the output to address that situation, and the learning objectives that may be involved.

As stated, this is a minimal example, and in practice further refinement would be required. For our production platform, the schema contains xxx different elements.

# Results and Discussion

The outcome of implementing this structured approach to scenario generation, using tailored prompts and data templates in conjunction with AI (specifically LLMs), has been profoundly positive. A system has been effectively established that enables the rapid creation of high-quality simulation scenarios. This system is characterized by its efficiency and flexibility, allowing for swift adaptation and customization of scenarios to meet diverse educational objectives and simulation requirements.

### Acknowledging the Limitations in Data Representation

The absence of qualification or quantification of results in this report should be acknowledged. This deliberate omission is attributed to several factors inherent in AI development, particularly in the context of online language models. The rapid development cycle, frequent updates, and opaque nature of these models pose significant challenges in producing data with assured quality and applicability.

This challenge is compounded by the wide variation in scope and complexity of medical scenarios, as well as the inherently subjective nature of assessing such material.

In the dynamic landscape of AI, where advancements and refinements occur at a rapid pace, any data collected or quantified may quickly become obsolete or irrelevant. Therefore, attempting to provide concrete data or results within the scope of this report would be impractical and potentially misleading.

Instead, the focus has been on providing insights, observations, and considerations based on current knowledge and understanding, recognizing the inherent limitations and uncertainties associated with the subject matter. Additionally, the worked example provided enables readers to rapidly assess the current state of platforms and tools without relying on potentially outdated or unrepresentative data.

### Enhanced Scenario Generation

Using custom prompts and semi-structured data templates has greatly streamlined the process of generating detailed and clinically relevant simulation scenarios. Guiding the AI with precise prompts at two levels—initial setup and within the template itself—has elicited focused and accurate content from LLM. This approach ensures that each scenario is grounded in the educational goals and clinical realities it aims to simulate.

### Flexibility and Customization

One of the standout features of this approach is the flexibility it offers. Educators can input specific subjects or details of the simulation they wish to create, tailoring the generated scenarios to the unique needs of their curriculum or training objectives. This level of customization is crucial for creating simulations that are not only relevant but also engaging for learners.

### Practical Applications

The practical applications of this system extend beyond the initial scenario creation. For instance, the data generated for scenario progression can be directly used to program manikin-based patient simulators as seen in Figure A3, enhancing the realism and effectiveness of the simulation training. This integration showcases the system's potential to not only create engaging narratives but also to provide concrete, practical tools for medical education.

### Exploring an API-Based Model

This section looks at the theoretical advantages and disadvantages of automation and scalability offered by current LLM application programing interfaces (APIs), including the feasibility of seamless integration with diverse educational platforms, and the potential for heightened customization and control over the scenario generation process. For those unfamiliar with APIs, an API is a set of rules and protocols that allows different software applications to communicate with each other. The following discussion is about the GPT3.5 and GPT4 APIs which provide a way for developers to interact with the model programmatically, sending inputs to the model and receiving outputs, such as generated text or responses.

### Automation and Scalability

An API-based model could automate the flow of data between this system and the AI, facilitating the generation of scenarios on a larger scale and with greater speed.

### Integration Flexibility

APIs offer the potential for seamless integration with a wider range of educational tools and platforms, expanding the utility and application of the generated scenarios across different learning environments.

### Customization and Control

With direct API access, there might be more opportunities to customize the AI's behavior and optimize the generation process for the specific needs, enhancing the quality and relevance of the scenarios.

### Challenges and Considerations of an API-Based Model

Maintaining and updating the context across interactions through an API can be complex, especially when dealing with intricate medical scenarios that require a nuanced understanding and continuity.

Developing and maintaining an API-based integration requires significant technical resources, including expertise in API management, data security, and system architecture.

As of the writing of this paper, GPT3.5 is freely available, and GPT4 can be accessed using a set subscription model. An API-based is charged based on the number of requests, computational resources consumed, or data transferred. These costs can generally be roughly calculated in advance but are still less predictable than set price models. These considerations need to be evaluated against the budget and projected usage to ensure affordability.

### Moving Forward

While this system is designed to be both accessible and straightforward, catering primarily to direct input and manipulation within platforms like ChatGPT, there is a growing consideration for the transition towards an API-based model for AI interactions. This approach could significantly enhance the efficiency and scalability of generating medical simulation scenarios but introduces its own set of challenges.

## Challenges

### Input Understanding and AI Interpretation:
One of the primary challenges faced was ensuring that the AI accurately interpreted the context and requirements of each simulation scenario. Despite using structured prompts and templates, there were instances where the AI struggled to understand the nuances of the input data, resulting in outputs that did not fully align with clinical realities or learning objectives. Manual review and adjustments were often necessary to correct these discrepancies. While these variances were generally minor and major errors were rare, maintaining simulation fidelity in the field of medicine requires careful consideration of input understanding and data interpretation.

### Output Accuracy and Error Handling:
In addition to input understanding, ensuring the accuracy of AI-generated output was equally challenging. Despite efforts to refine the model and provide clear guidelines, there were instances where the AI produced outputs that deviated from expected clinical or educational standards. Robust mechanisms for data validation and error correction were essential to address these issues.

### User Interface and Experience

Developing an intuitive and user-friendly interface emerged as a critical consideration. The goal was to enable educators to seamlessly input scenario requirements and interpret the AI-generated content without undue complexity. Achieving a design that catered to users across a spectrum of familiarity with AI and simulation technologies required iterative development and focused attention on user experience.

### Adapting to the Continuous Evolution of AI

A significant challenge in leveraging platforms like ChatGPT for generating medical simulation scenarios is their inherently dynamic nature. These platforms are continuously updated and refined, with models evolving to offer better accuracy, prompt interpretation, and output quality. While these updates are generally beneficial, they can occasionally disrupt established workflows or templates that were optimized for previous versions of the model.

## Conclusion

This paper introduces a novel approach to the development of medical simulation scenarios through the integration of semi-structured data and Large Language Models (LLMs), such as OpenAI's ChatGPT3.5. By automating the scenario creation process, this method significantly reduces the time and effort traditionally required, allowing for the production of a wider variety of simulation scenarios that are tailored to specific learning needs. The resulting increase in efficiency not only enhances the adaptability and realism of these simulations but also supports rapid iteration and customization. This helps to contribute to the dynamic and ever-evolving demands of medical education.

Feedback from educators has been positive, specifically noting the time saving nature of the technology. The use of semi-structured data ensures that the scenarios are clinically relevant and educationally valuable, while the automation provided by LLMs allows educational programs to expand their offerings without the corresponding increase in resource expenditure.

The success of this approach underscores the potential of combining AI with semi-structured data to transform medical education. It highlights an easily scalable and dynamic solution that could greatly impact the development and application of simulation scenarios, ultimately improving learning outcomes and patient care standards. However, the necessity for oversight by subject matter experts remains, to ensure the clinical accuracy and educational value of the AI-generated content. This balanced approach maximizes the benefits of AI while maintaining the essential human element crucial to medical education.

# Figure Legend:

Figure 1: Automated Scenario Generation Process Flow

Illustrates the automated process from user input through AI processing to the final scenario generation. Highlights include the transformation of clinical scenario ideas into structured prompts for ChatGPT, subsequent JSON data generation, and the compilation of validated scenario data into educational documents.

Figure A1: Application Interface Overview

Displays the application's interface, focusing on the user input box for scenario generation and the subsequent steps leading to the generation of JSON text by ChatGPT, emphasizing the initial step in scenario creation.

Figure A3: Laerdal Manikin Scenario Integration

Depicts the process of converting generated scenario data into a Laerdal manikin scenario file, highlighting how the application preprograms simulation phases and vital signs, showcasing the direct application of generated scenarios in simulation-based education.

# Appendix A
## Our Implementation Example

Figure A1 shows the main screen of the application. It has a box for the user to input their scenario idea or clinical need. "Copy General Info Schema" is then clicked on which embeds the text that the user input into a question prompt for ChatGPT, and then has the schema appended to it. The user then pastes the combined prompt into ChatGPT's input. After clicking send, the GPT will then begin to output the completed JSON text. The user would then copy this data to the clipboard, and press paste under "General Info". This will check the data, make any automatic corrections that it can, and then display the output on the right side in the PDF pane.

The user would then follow the same process for the scenario progression part of the schema. The software automatically assembles and formats the data from both schemas into the final document. Figure A2 shows an example of the completed scenario which can be printed. Currently, JSON can either be edited directly in a text editor or rewritten based on requests to ChatGPT at creation time. There are plans to alter the program to allow inline editing and more granular control over the input/output.

Once the user is satisfied with the scenario that they've created, if they have a Laerdal manikin, they can click on "Output Laerdal Scene" which will create a scenario file that can be opened in the Laerdal Instructor Application. Figure A3 shows that this has created each of the phases in the software and preprograms the vitals for each phase.

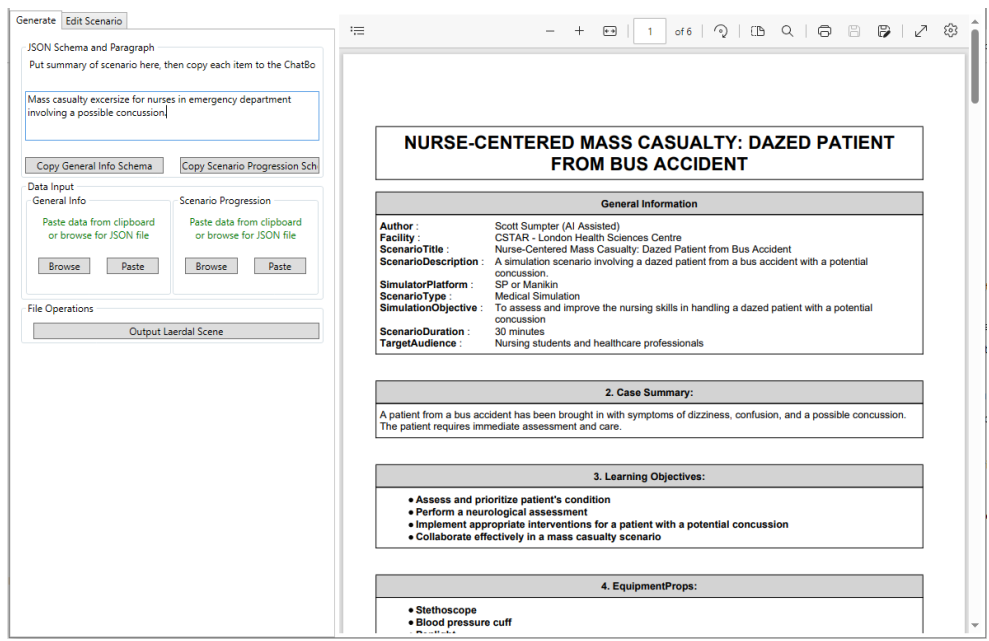

*Figure A1 : Application Interface Overview*

# NURSE-CENTERED MASS CASUALTY: DAZED PATIENT FROM BUS ACCIDENT

## General Information

**Author** : Scott Sumpter (AI Assisted)
**Facility** : CSTAR - London Health Sciences Centre
**ScenarioTitle** : Nurse-Centered Mass Casualty: Dazed Patient from Bus Accident
**ScenarioDescription** : A simulation scenario involving a dazed patient from a bus accident with a potential concussion.
**SimulatorPlatform** : SP or Manikin
**ScenarioType** : Medical Simulation
**SimulationObjective** : To assess and improve the nursing skills in handling a dazed patient with a potential concussion
**ScenarioDuration** : 30 minutes
**TargetAudience** : Nursing students and healthcare professionals

## 2. Case Summary:

A patient from a bus accident has been brought in with symptoms of dizziness, confusion, and a possible concussion. The patient requires immediate assessment and care.

## 3. Learning Objectives:

- **Assess and prioritize patient's condition**
- **Perform a neurological assessment**
- **Implement appropriate interventions for a patient with a potential concussion**
- **Collaborate effectively in a mass casualty scenario**

## 4. EquipmentProps:

- **Stethoscope**
- **Blood pressure cuff**
- **Penlight**
- **Gloves**
- **Cervical collar**
- **Spine board**
- **IV supplies**

## 5. Environment:

Emergency department with multiple patients from the bus accident, including paramedics and other healthcare professionals.

| 6. Case Presentation: |
|---|
| The patient is a 35-year-old male who was on a bus involved in a collision. On arrival, he appears dazed and disoriented. His vital signs are as follows:<br>- Heart rate: 90 bpm<br>- Blood pressure: 130/80 mm Hg<br>- Respiratory rate: 18 breaths/min<br>- Temperature: 98.6°F (37°C)<br><br>No tests have been performed yet. The medical background includes information about the bus accident and the mechanism of injury. |

| 8. Debriefing Points: |
|---|
| - **Assessment of the patient's level of consciousness**<br>- **Neurological assessment findings**<br>- **Prioritization of care in a mass casualty situation**<br>- **Effective communication and teamwork** |

| 9. Lab Results: | | |
|---|---|---|
| Test | Results | Normal |
| CT Head Scan | No acute intracranial abnormalities detected. | Normal |
| CBC (Complete Blood Count) | Within normal limits | Normal |

| Scenario Progression |||
|---|---|---|
| Phase | Learner Actions | Modifiers & Triggers |
| **1) Initial Assessment**<br>**BP: 130/80 mm Hg**<br>**HR: 90 bpm**<br>**RR: 18 breaths/min**<br>**SpO2: 95%**<br>**Temp: 37°C**<br>**Rhythm: SINUS**<br>**GCS:**<br>**E: 4 (Opens eyes in response to voice)**<br>**V: 4 (Confused)**<br>**M: 6 (Obeys commands)**<br>**Other: Patient appears dazed and confused.** | Perform initial patient assessment<br>Assess neurological status<br>Administer oxygen if necessary | **Modifiers:**<br>Modifier: Administered oxygen<br>Result: SpO2 increased to 98%<br><br>**Triggers:**<br>Trigger: Completion of initial assessment<br>Result: Transition to Phase 2 |
| **2) Secondary Assessment**<br>**BP: 120/80 mm Hg**<br>**HR: 95 bpm**<br>**RR: 20 breaths/min**<br>**SpO2: 98%**<br>**Temp: 37°C**<br>**Rhythm: SINUS**<br>**GCS:**<br>**E: 3 (Opens eyes in response to voice)**<br>**V: 3 (Inappropriate words)**<br>**M: 5 (Localizes to pain)**<br>**Other: Patient's condition is stable.** | Perform detailed neurological examination<br>Apply cervical collar<br>Prepare for imaging tests | **Modifiers:**<br>Modifier: Applied cervical collar<br>Result: Patient reports decreased neck discomfort<br><br>**Triggers:**<br>Trigger: Completion of secondary assessment<br>Result: Transition to Phase 3 |
| **3) Deterioration**<br>**BP: 90/60 mm Hg**<br>**HR: 110 bpm**<br>**RR: 22 breaths/min**<br>**SpO2: 92%**<br>**Temp: 37°C**<br>**Rhythm: SINUS**<br>**GCS:**<br>**E: 2 (Opens eyes to pain)**<br>**V: 2 (Incomprehensible sounds)**<br>**M: 4 (Withdraws to pain)**<br>**Other: Patient shows signs of deterioration.** | Recognize signs of clinical deterioration<br>Administer IV fluids immediately<br>Reassess patient condition and prepare for possible escalation of care | **Modifiers:**<br>Modifier: Administered IV fluids<br>Result: BP temporarily stabilized to 100/70 mm Hg<br><br>**Triggers:**<br>Trigger: Observation of further deterioration or stabilization<br>Result: Transition to Phase 4 |
| **4) Emergency Intervention**<br>**BP: 100/70 mm Hg**<br>**HR: 115 bpm**<br>**RR: 24 breaths/min**<br>**SpO2: 90%**<br>**Temp: 37°C**<br>**Rhythm: SINUS**<br>**GCS:**<br>**E: 1 (Does not open eyes)**<br>**V: 1 (Makes no sounds)**<br>**M: 3 (Decerebrate posturing)**<br>**Other: Patient is now unresponsive, indicating a need for urgent intervention.** | Initiate advanced airway management<br>Call for emergency neurological consultation<br>Prepare for immediate transport to imaging | **Modifiers:**<br>Modifier: Administered advanced airway management<br>Result: SpO2 improved to 94%<br><br>**Triggers:**<br>Trigger: Successful airway management<br>Result: Transition to Phase 5 |
| **5) Conclusion and Debriefing**<br>**BP: 120/80 mm Hg**<br>**HR: 95 bpm**<br>**RR: 20 breaths/min**<br>**SpO2: 94%**<br>**Temp: 37°C**<br>**Rhythm: SINUS**<br>**GCS:**<br>**E: 4 (Opens eyes in response to voice)**<br>**V: 4 (Confused)**<br>**M: 6 (Obeys commands)**<br>**Other: Patient is stable post-intervention and awaiting further diagnostic evaluation.** | Review the case and interventions<br>Discuss the outcome and potential improvements<br>Debrief on teamwork and communication throughout the scenario | **Modifiers:** |

| SP Instructions |  |
|---|---|
| **Character**<br>Simulated Patient (SP)<br><br>**Summary**<br>The SP is a 35-year-old male who was on the bus during the accident. He is confused and disoriented, experiencing dizziness. He will respond to questions but may not remember all the details of the accident.<br><br>**Moulage**<br>Apply moulage to simulate bruising on the forehead and a possible minor head injury. | |
| **Phase 1 Instructions** | |
| **Description**<br>Patient is dazed but responsive, appearing confused and disoriented.<br><br>**Physical Instructions**<br>Act dazed, confused, and disoriented, responding slowly to verbal stimuli.<br><br>**Vocal Emotional Instructions**<br>Mumble incoherently when spoken to, showing signs of confusion. | **Possible Question:** Can you tell me your name?<br>**Response:** Responds with hesitation or incorrect answers.<br><br>**Possible Question:** Do you know where you are?<br>**Response:** Shows confusion or incorrect perception of the environment. |
| **Phase 2 Instructions** | |
| **Description**<br>Patient remains dazed but shows slight improvement in responsiveness.<br><br>**Physical Instructions**<br>Act with slight improvement, responding more consistently to verbal stimuli.<br><br>**Vocal Emotional Instructions**<br>Use confused but slightly more coherent speech. | **Possible Question:** Can you tell me how you're feeling?<br>**Response:** Expresses discomfort or confusion verbally but more coherently. |
| **Phase 3 Instructions** | |
| **Description**<br>Patient's condition deteriorates, becoming less responsive.<br><br>**Physical Instructions**<br>Show signs of significant deterioration, being less responsive to stimuli.<br><br>**Vocal Emotional Instructions**<br>Make incomprehensible sounds, no coherent speech. | **Possible Question:** Can you hear me?<br>**Response:** Does not respond or responds with non-specific movement. |
| **Phase 4 Instructions** | |

| **Description**<br>Patient requires immediate emergency intervention, unresponsive.<br><br>**Physical Instructions**<br>Remain unresponsive to all external stimuli.<br><br>**Vocal Emotional Instructions**<br>No vocal sounds, indicating severe deterioration. | **Possible Question:** Squeeze my hand if you can hear me.<br>**Response:** No response, indicating lack of consciousness. |
|---|---|
| **Phase 5 Instructions** ||
| **Description**<br>Scenario concludes with the patient stabilized and prepared for further treatment.<br><br>**Physical Instructions**<br>No additional instructions, as the scenario focuses on debriefing.<br><br>**Vocal Emotional Instructions**<br>No additional instructions, as the scenario focuses on debriefing. | |

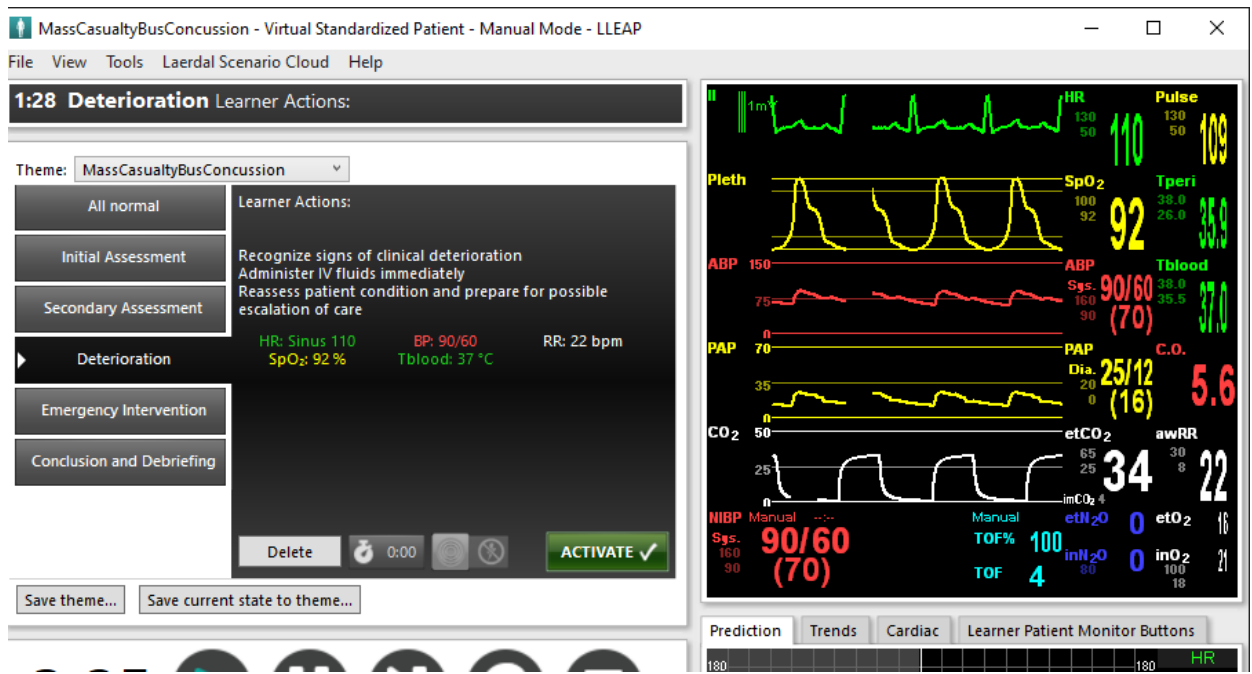

*Figure A2: Laerdal Manikin Scenario Integration*

# Appendix B
## Simplified example

The following two prompts were pasted individually into chatgpt, the response from chatgpt of each exchange is provided after the prompt.

**Prompt:**
```
Adhering strictly to the following schema, fill in the values to create a complete medical simulation. Output
should be the JSON object without any schema elements. The simulation will be about a myocardial infarction:
{
  "scenario": {
    "type": "object",
    "properties": {
      "generalinfo": {
        "$id": "#root/scenario/generalinfo",
        "type": "object",
        "properties": {
          "casepresentation": {
            "$id": "#root/scenario/generalinfo/casepresentation",
            "description": "The initial presentation of the patient in the scenario, including relevant
clinical details. Should include a readout of key vitals as well as if any tests were performed or any other
inbound issues occurred. Medical background should be provided if it would be available.",
            "type": "string"
          },
          "scenariotitle": {
            "$id": "#root/scenario/generalinfo/scenariotitle",
            "type": "string"
          },
          "casesummary": {
            "$id": "#root/scenario/generalinfo/casesummary",
            "description": "A brief summary of the scenario and the patient's condition.",
            "type": "string"
          },
          "debriefingpoints": {
            "$id": "#root/scenario/generalinfo/debriefingpoints",
            "type": "object",
            "properties": {
              "point": {
                "$id": "#root/scenario/generalinfo/debriefingpoints/point",
                "description": "Key discussion points for the debriefing session following the scenario.",
                "type": "array",
                "items": {
                  "type": "string"
                }
              }
            }
          },
          "learningobjectives": {
            "$id": "#root/scenario/generalinfo/learningobjectives",
            "type": "object",
            "properties": {
              "objective": {
                "$id": "#root/scenario/generalinfo/learningobjectives/objective",
                "description": "The learning objectives for the scenario highlighting the specific knowledge or
skills to be acquired.",
                "items": {
                  "type": "string"
                },
                "type": "array"
              }
            }
          }
        }
      }
    }
  }
}
```

**Prompt 1 response:**

```json
{
  "generalinfo": {
    "casepresentation": "A 52-year-old male presents to the emergency department with severe chest pain radiating to his left arm, lasting for more than 30 minutes. The patient describes the pain as a pressing sensation on his chest. Key vitals upon presentation: blood pressure 140/90 mmHg, heart rate 110 BPM, respiratory rate 20 breaths/min, oxygen saturation 94% on room air. Past medical history includes hypertension and type 2 diabetes mellitus. The patient is a smoker with a 30-pack-year history.",
    "scenariotitle": "Acute myocardial infarction management",
    "casesummary": "This scenario involves a 52-year-old male with a history of hypertension and type 2 diabetes, presenting with symptoms suggestive of an acute myocardial infarction (ami).",
    "debriefingpoints": {
      "point": [
        "Recognition and immediate management of acute myocardial infarction",
        "Importance of rapid triage and ECG interpretation in suspected AMI cases",
        "Discussion on the use of thrombolytics vs. PCI (percutaneous coronary intervention) depending on hospital capabilities and time frames",
        "Management of risk factors and secondary prevention post-AMI"
      ]
    },
    "learningobjectives": {
      "objective": [
        "Identify the signs and symptoms of an acute myocardial infarction",
        "Understand the initial emergency management steps for a patient presenting with AMI",
        "Describe the criteria for thrombolytic therapy versus PCI",
        "Explain the importance of risk factor management in preventing future cardiac events"
      ]
    }
  }
}
```

**Prompt 2:**
Based on the general information context established above, fill in the values to create a complete medical simulation. Provide at least 3 to 5 scenario progression steps. Scenario progression should have a clear beginning, middle and end.  Beginning should be initial assessment.  Output should be the JSON object without any schema elements. The simulation will be about a myocardial infarction:

```
{
  "scenario": {
    "type": "object",
    "properties": {
      "scenarioprogression": {
        "$id": "#root/scenario/scenarioprogression",
        "type": "object",
        "properties": {
          "steps": {
            "$id": "#root/scenario/scenarioprogression/steps",
            "type": "array",
            "items": {
              "type": "object",
              "properties": {
                "stepnumber": {
                  "$id": "#root/scenario/scenarioprogression/steps/stepnumber",
                  "type": "integer",
                  "description": "The order of the step within the scenario progression."
                },
                "patientstatus": {
                  "$id": "#root/scenario/scenarioprogression/steps/patientstatus",
                  "type": "string",
                  "description": "A brief description of the patient's current condition or changes in condition at this step."
                },
                "vitals": {
                  "$id": "#root/scenario/scenarioprogression/steps/vitals",
                  "type": "object",
                  "properties": {
                    "HR": {
                      "$id": "#root/scenario/scenarioprogression/steps/vitals/HR",
                      "type": "integer",
                      "description": "Heart rate in beats per minute."
                    },
                    "BP": {
                      "$id": "#root/scenario/scenarioprogression/steps/vitals/BP",
                      "type": "string",
                      "description": "Blood pressure in mmHg, formatted as systolic/diastolic."
                    },
                    "SpO2": {
                      "$id": "#root/scenario/scenarioprogression/steps/vitals/SpO2",
                      "type": "integer",
                      "description": "Oxygen saturation percentage."
                    }
                  },
                  "description": "The vital signs of the patient at this step of the scenario."
                }
              }
            }
          }
        }
      }
    }
  }
}
```

**Prompt 2 response:**

```
{
  "scenarioprogression": {
    "steps": [
      {
        "stepnumber": 1,
        "patientstatus": "Upon arrival, the patient is in distress, complaining of severe chest pain and shortness of breath.",
        "vitals": {
          "HR": 110,
          "BP": "140/90",
          "SpO2": 94
        }
      },
      {
        "stepnumber": 2,
        "patientstatus": "After administering oxygen and aspirin, the patient's pain persists. An ecg is performed, showing st-segment elevation in leads ii, iii, and avf.",
        "vitals": {
          "HR": 120,
          "BP": "135/85",
          "SpO2": 92
        }
      },
      {
        "stepnumber": 3,
        "patientstatus": "The patient's condition suddenly deteriorates, leading to cardiac arrest. Cpr is immediately initiated, and advanced cardiac life support protocols are followed.",
        "vitals": {
          "HR": 0, // indicative of cardiac arrest
          "BP": "0/0", // not measurable during cardiac arrest
          "SpO2": 85 // falling due to lack of circulation
        }
      },
      {
        "stepnumber": 4,
        "patientstatus": "After three rounds of CPR and one shock from the defibrillator, a pulse is regained. The patient is intubated and rushed for emergency percutaneous coronary intervention (PCI).",
        "vitals": {
          "HR": 90, // post-resuscitation
          "BP": "110/70", // post-resuscitation
          "SpO2": 95 // on mechanical ventilation
        }
      },
      {
        "stepnumber": 5,
        "patientstatus": "post-PCI, the patient's condition stabilizes significantly. The patient remains intubated but shows signs of recovery. Continuous monitoring in the ICU is maintained.",
        "vitals": {
          "HR": 80,
          "BP": "120/80",
          "SpO2": 98
        }
      },
      {
        "stepnumber": 6,
        "patientstatus": "The patient is successfully extubated and demonstrates significant improvement. Discharged with strict follow-up for cardiac rehabilitation, medication management, and lifestyle modification counseling.",
        "vitals": {
          "HR": 70,
          "BP": "115/75",
          "SpO2": 98
        }
      }
    ]
  }
}
```

**Python Example:**

```python
general_info = """ PASTE GENERAL INFO HERE """
scenario_progression = """ PASTE SCENARIO PROGRESSION HERE """
# Display General Information
general_info = json.loads(general_info)
scenario_progression = json.loads(scenario_progression)

print("General Information:")
print("Scenario Title:", general_info['generalinfo']['scenariotitle'])
print("Case Presentation:", general_info['generalinfo']['casepresentation'])
print("Case Summary:", general_info['generalinfo']['casesummary'])

print("\nDebriefing Points:")
for point in general_info['generalinfo']['debriefingpoints']['point']:
    print("-", point)

print("\nLearning Objectives:")
for objective in general_info['generalinfo']['learningobjectives']['objective']:
    print("-", objective)

# Display Scenario Progression
print("\nScenario Progression:")
for step in scenario_progression['scenarioprogression']['steps']:
    print(f"\nStep {step['stepnumber']}:")
    print("Patient Status:", step['patientstatus'])
    print("Vitals:")
    print("  Heart Rate:", step['vitals']['HR'])
    print("  Blood Pressure:", step['vitals']['BP'])
    print("  Oxygen Saturation:", step['vitals']['SpO2'])
```

**Python Output:**

```
Output:
General Information:
Scenario Title: Acute myocardial infarction management
Case Presentation: A 52-year-old male presents to the emergency department with severe chest pain radiating to his left arm,
lasting for more than 30 minutes. The patient describes the pain as a pressing sensation on his chest. Key vitals upon
presentation: blood pressure 140/90 mmHg, heart rate 110 BPM, respiratory rate 20 breaths/min, oxygen saturation 94% on room
air. Past medical history includes hypertension and type 2 diabetes mellitus. The patient is a smoker with a 30-pack-year
history.
Case Summary: This scenario involves a 52-year-old male with a history of hypertension and type 2 diabetes, presenting with
symptoms suggestive of an acute myocardial infarction (AMI).

Debriefing Points:
- Recognition and immediate management of acute myocardial infarction
- Importance of rapid triage and ECG interpretation in suspected AMI cases
- Discussion on the use of thrombolytics vs. PCI (percutaneous coronary intervention) depending on hospital capabilities and
time frames
- Management of risk factors and secondary prevention post-AMI

Learning Objectives:
- Identify the signs and symptoms of an acute myocardial infarction
- Understand the initial emergency management steps for a patient presenting with AMI
- Describe the criteria for thrombolytic therapy versus PCI
- Explain the importance of risk factor management in preventing future cardiac events

Scenario Progression:

Step 1:
Patient Status: Upon arrival, the patient is in distress, complaining of severe chest pain and shortness of breath.
Vitals:
   Heart Rate: 110
   Blood Pressure: 140/90
   Oxygen Saturation: 94

Step 2:
Patient Status: After administering oxygen and aspirin, the patient's pain persists. An ECG is performed, showing ST-segment
elevation in leads II, III, and AVF.
Vitals:
   Heart Rate: 120
   Blood Pressure: 135/85
   Oxygen Saturation: 92

Step 3:
Patient Status: The patient's condition suddenly deteriorates, leading to cardiac arrest. CPR is immediately initiated, and
advanced cardiac life support protocols are followed.
Vitals:
   Heart Rate: 0
   Blood Pressure: 0/0
   Oxygen Saturation: 85

Step 4:
Patient Status: After three rounds of CPR and one shock from the defibrillator, a pulse is regained. The patient is intubated
and rushed for emergency percutaneous coronary intervention (PCI).
Vitals:
   Heart Rate: 90
   Blood Pressure: 110/70
   Oxygen Saturation: 95

Step 5:
Patient Status: Post-PCI, the patient's condition stabilizes significantly. The patient remains intubated but shows signs of
recovery. Continuous monitoring in the ICU is maintained.
Vitals:
   Heart Rate: 80
   Blood Pressure: 120/80
   Oxygen Saturation: 98

Step 6:
Patient Status: The patient is successfully extubated and demonstrates significant improvement. Discharged with strict follow-
up for cardiac rehabilitation, medication management, and lifestyle modification counseling.
Vitals:
   Heart Rate: 70
   Blood Pressure: 115/75
   Oxygen Saturation: 98
```

*Table A1 Python Code Output for Example JSON*